\documentclass[letterpaper]{article}    
\usepackage{aaai2026}                 

\usepackage{times}                      
\usepackage{helvet}                     
\usepackage[hyphens]{url}               
\usepackage{graphicx}                   
\usepackage{natbib}     
\usepackage{caption}    
\usepackage{amsfonts,amsmath,amssymb,amsthm}
\usepackage{booktabs}
\usepackage{threeparttable}
\usepackage{multirow}
\usepackage{bm}

\usepackage{color}
\usepackage{colortbl}
\usepackage{xcolor}
\usepackage{tcolorbox}

\makeatletter
\newif\if@restonecol
\makeatother

\usepackage[linesnumbered,ruled,vlined]{algorithm2e}
\usepackage{algpseudocode}

\SetKwComment{Comment}{$\triangleright$\ }{}

\DeclareMathOperator*{\argmax}{arg\,max}

\renewcommand{\paragraph}[1]{\smallskip\noindent\textbf{#1.}}
\renewcommand{\subparagraph}[1]{\smallskip\noindent\textbf{\underline{#1.}}}

\newtcolorbox{graybox}{
  colback=gray!10,
  colframe=gray!50,
  boxrule=1pt,
  arc=3pt,
  left=5pt,
  right=5pt,
  top=2pt,
  bottom=0pt
}

\title{Knowledge Completes the Vision: A Multimodal Entity-aware Retrieval-Augmented Generation Framework for News Image Captioning}

\author{
  Xiaoxing~You\textsuperscript{\rm 1,}\thanks{Equal contribution.},\;
  Qiang~Huang\textsuperscript{\rm 2,}\footnotemark[1],\;
  Lingyu~Li\textsuperscript{\rm 1},\;
  Chi~Zhang\textsuperscript{\rm 3},\;
  Xiaopeng~Liu\textsuperscript{\rm 3},\;
  Min~Zhang\textsuperscript{\rm 2},\;
  Jun~Yu\textsuperscript{\rm 2,4,}\thanks{Corresponding author.}
}
\affiliations{
  \textsuperscript{\rm 1}Hangzhou Dianzi University, Hangzhou, China \\ 
  \textsuperscript{\rm 2}Harbin Institute of Technology (Shenzhen), Shenzhen, China \\
  \textsuperscript{\rm 3}People's Daily, Beijing, China \\ 
  \textsuperscript{\rm 4}Peng Cheng Laboratory, Shenzhen, China \\
  \{youxiaoxing,~lilingyu0571\}@hdu.edu.cn, \{huangqiang,~zhangmin2021,~yujun\}@hit.edu.cn, \{zhangchi,~liuxiaopeng\}@pdnews.cn
}

\begin{document}
\maketitle

\begin{abstract}
News image captioning aims to produce journalistically informative descriptions by combining visual content with contextual cues from associated articles. 
Despite recent advances, existing methods struggle with three key challenges: (1) incomplete information coverage, (2) weak cross-modal alignment, and (3) suboptimal visual-entity grounding.
To address these issues, we introduce \textbf{MERGE}, the first \textbf{M}ultimodal \textbf{E}ntity-aware \textbf{R}etrieval-augmented \textbf{GE}neration framework for news image captioning. 
MERGE constructs an entity-centric multimodal knowledge base (EMKB) that integrates textual, visual, and structured knowledge, enabling enriched background retrieval. 
It improves cross-modal alignment through a multistage hypothesis-caption strategy and enhances visual-entity matching via dynamic retrieval guided by image content.
Extensive experiments on GoodNews and NYTimes800k show that MERGE significantly outperforms state-of-the-art baselines, with CIDEr gains of +6.84 and +1.16 in caption quality, and F1-score improvements of +4.14 and +2.64 in named entity recognition.
Notably, MERGE also generalizes well to the unseen Visual News dataset, achieving +20.17 in CIDEr and +6.22 in F1-score, demonstrating strong robustness and domain adaptability.
\end{abstract}

\begin{links}
    \link{Code}{https://github.com/youxiaoxing/MERGE}
\end{links}


\section{Introduction}
\label{sec:intro}

News articles typically include images accompanied by captions that blend visual elements with contextual details, enhancing reader comprehension and engagement.
Unlike vanilla image captioning methods \citep{vinyals2016show, hossain2019comprehensive, yu2019multimodal, xu2023versatile}, which primarily describe visible content, news image captioning demands both precise entity recognition and the incorporation of deeper contextual knowledge.
Editors must analyze key elements--such as people, events, time, and location--and craft captions tailored to diverse journalistic contexts, where the same image may require entirely different descriptions \citep{nguyen2023show}.

\begin{figure}[t]
  \centering
  \includegraphics[width=0.99\columnwidth]{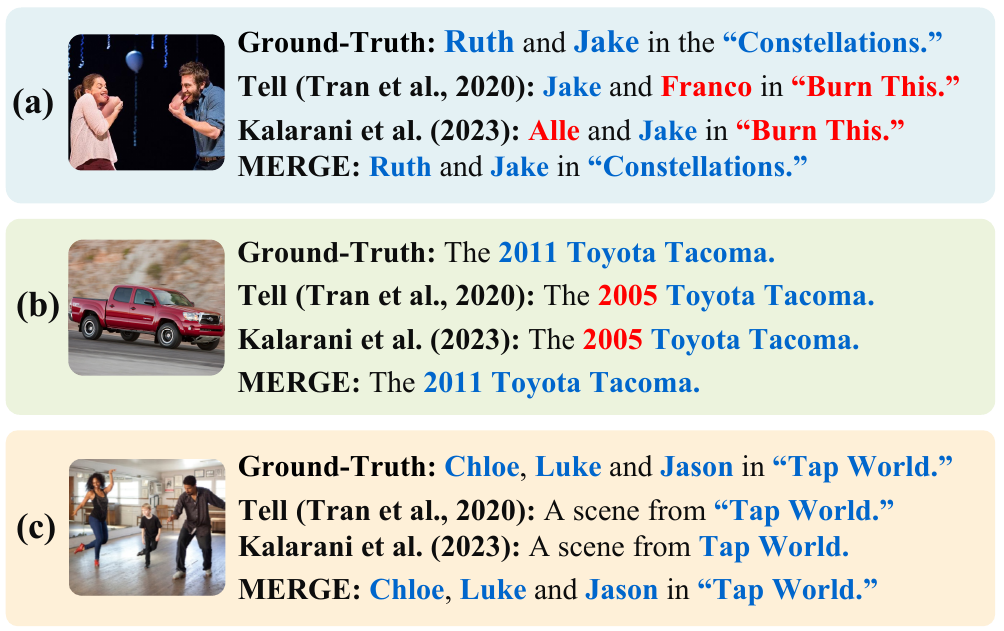}
  \caption{Challenges in news image captioning: (a) Identifying entities absent from the article; (b) Aligning numerical details and visual objects across modalities; (c) Disambiguating entities in images with multiple subjects.}
  \label{fig:motivating_example}
\end{figure}

Automated news image captioning has been widely studied to assist editors.
Early template-based systems~\citep{Ramisa2018Break, biten2019good, hu2020icecap} filled predefined templates with entities.
While effective for structured output, these methods often yield rigid captions lacking nuanced context.
Transformer-based models~\citep{tran2020transform, zhao2024boosting} have introduced richer modeling of visual features--such as faces and objects--to improve entity-aware captioning. 
However, they often struggle to extract precise details from long or noisy articles, leading to incomplete or generic captions.
Another prominent line of work~\citep{zhou2022focus, qu2024visually} focuses on extracting relevant textual contexts from articles. 
Techniques in this direction often leverage pre-trained or fine-tuned CLIP models~\citep{radford2021clip} to retrieve salient sentences while minimizing redundancy. 
Yet, these methods typically fall short in establishing deep semantic connections between visual elements and textual narratives.
More recently, Multimodal Large Language Models (MLLMs)~\citep{xu2024cross, zhang2024entity} have shown great promise by jointly modeling visual and textual modalities. 
Their advanced reasoning capabilities and flexibility make them well-suited for the complex demands of news image captioning.

Despite significant progress, as illustrated in Figure \ref{fig:motivating_example}, existing approaches continue to face three critical challenges:
\begin{itemize}
  \item \textbf{Incomplete Information Coverage:} 
  Generating accurate captions often requires referencing entities not explicitly mentioned in the accompanying article.
  For example, as shown in Figure \ref{fig:motivating_example}(a), existing models fail to identify \texttt{Ruth Wilson} because she is absent from the text. 
  Current methods \citep{biten2019good, tran2020transform, xu2024cross, zhang2024entity} struggle to retrieve missing information and effectively integrate external knowledge \citep{xu2024understand}.
  
  \item \textbf{Limited Cross-modal Alignment:}
  At the sentence level, existing methods typically focus on either describing visual scenes~\citep{hu2020icecap, qu2024visually} or extracting entity-rich sentences~\citep{zhou2022focus, zhang2024entity}, but struggle to holistically align visual objects with numerical details--for example, linking the \texttt{Toyota Tacoma} to its \texttt{2011} launch year, as depicted in Figure \ref{fig:motivating_example}(b).

  \item \textbf{Suboptimal Visual-Entity Grounding:}
  Linking visual cues to correct named entities remains challenging, especially in images with multiple people or objects (Figure \ref{fig:motivating_example}(c)). 
  Implicit grounding methods \citep{qu2024visually} offer limited control, while fine-tuned models \citep{zhao2024boosting} struggle with unseen entities.
  Recent Retrieval-Augmented Generation (RAG) approaches \citep{xu2024understand} use resources like ConceptNet \citep{speer2017conceptnet} for entity understanding but still lack robust visual-text integration, highlighting the need for multimodal RAG frameworks.
\end{itemize}

To address these challenges, we propose \textbf{MERGE}, the first \textbf{M}ultimodal \textbf{E}ntity-aware \textbf{R}etrieval-augmented \textbf{GE}neration framework customized for news image captioning. 
By seamlessly integrating explicit multimodal knowledge with the implicit reasoning capabilities of MLLMs, MERGE introduces three key innovations:
\begin{itemize}
  \item \textbf{Information Enhancement:} 
  MERGE builds an Entity-centric Multimodal Knowledge Base (EMKB) that consolidates named entities, images, and structured background knowledge.
  This resource enables the model to supplement missing details absent from the article--for instance, identifying the actress \texttt{Ruth Wilson}, who might otherwise be overlooked (Figure \ref{fig:motivating_example}(a)).

  \item \textbf{Fine-grained Cross-modal Alignment:}
  To improve sentence-level alignment, MERGE introduces Hypothesis Caption-guided Multimodal Alignment (HCMA), which employs a three-stage Chain-of-Thought (CoT) prompting mechanism.
  This structured reasoning process enables accurate matching between visual cues and textual details, including nuances like the \texttt{2011} launch year of the \texttt{Toyota Tacoma} (Figure \ref{fig:motivating_example}(b)).

  \item \textbf{Precise Visual-Entity Alignment:} 
  For robust entity-level grounding, MERGE incorporates Retrieval-driven Multimodal Knowledge Integration (RMKI), which dynamically retrieves multimodal evidence and constructs background knowledge graphs from EMKB. 
  This allows MERGE to distinguish visually similar individuals and keep precise entity associations--for example, correctly identifying \texttt{Chloe}, \texttt{Luke}, and \texttt{Jason} in Figure~\ref{fig:motivating_example}(c).
\end{itemize}

We conduct extensive evaluations of MERGE on three real-world datasets: GoodNews, NYTimes800k, and Visual News.
For the GoodNews and NYTimes800k datasets, MERGE achieves new state-of-the-art performance, improving CIDEr by +6.84 and +1.16, and boosting named entity recognition F1-scores by +4.14 and +2.64, respectively.
Importantly, MERGE generalizes well to Visual News, outperforming prior methods with a +20.17 boost in CIDEr and +6.22 in F1-score, despite the dataset being excluded from EMKB construction.
Ablation studies further validate MERGE's architecture: HCMA enhances cross-modal alignment, while RMKI and EMKB significantly improve both caption quality and entity-level precision.
These results underscore MERGE's effectiveness in generating accurate, contextually grounded, and journalistically informative captions across diverse domains.


\section{Related Work}
\label{sec:related_work}

News image captioning aims to generate captions by integrating visual and textual information to produce contextually relevant descriptions, which can be broadly divided into three categories: (1) processing original articles, (2) extracting relevant contexts, and (3) incorporating MLLMs.

\paragraph{Methods Processing Original Articles}
These methods~\citep{biten2019good, yang2020image, zhang2022fine, zhang2023exploring, nguyen2023show, ajankar2024image, xu2024understand} use original news articles as inputs, often truncated due to model input constraints~\citep{zhou2022focus}.
\citet{biten2019good} introduced a two-stage template-based method using entity placeholders.
\citet{tran2020transform} proposed a transformer-based framework incorporating face and object embeddings for end-to-end captioning. 
\citet{yang2021journalistic} structured captions using six journalistic components, while \citet{liu2021visual} developed the Visual News dataset to enhance image-article connections.
\citet{zhang2022fine} combined CLIP with BART~\citep{lewis2020bart} to improve named entity recognition, and \citet{kalarani2023let} pre-trained OFA for tasks like visual entailment and keyword extraction.
\citet{zhao2024boosting} introduced an entity-matching module to build multimodal graphs linking faces, objects, and entities.
\citet{xu2024rule} applied prefix-tuning to inject news-specific semantic rules into BART for guided captioning.
MERGE advances this line of work by retrieving entities and background knowledge from a multimodal knowledge base, bridging visual cues with entities and dynamically handling new entities without additional fine-tuning.

\begin{figure*}[t]
  \centering
  \includegraphics[width=0.99\textwidth]{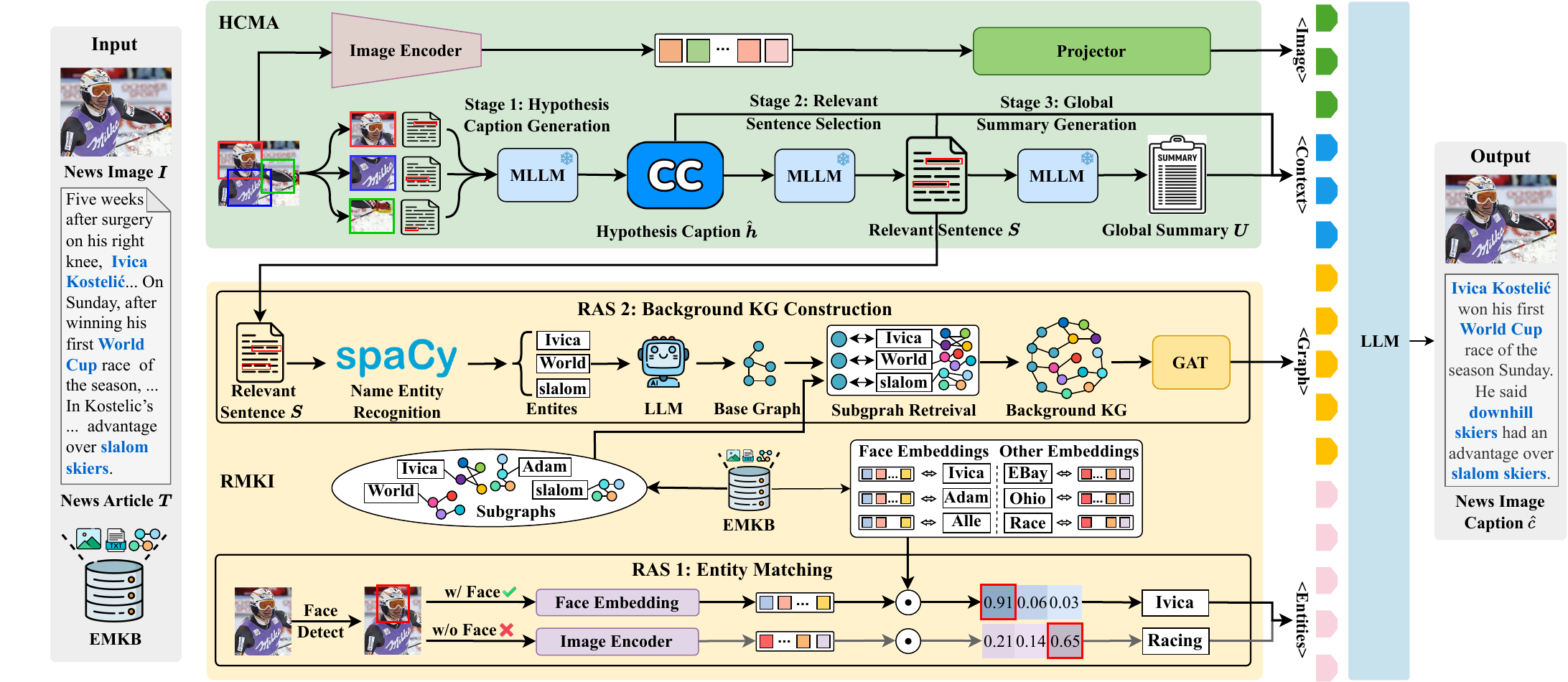}
  \caption{Overview of the MERGE framework.}
  \label{fig:model_overview}
\end{figure*}

\paragraph{Methods Using Extracted Relevant Contexts}
To reduce irrelevant information, these methods employ pre-trained or fine-tuned models like CLIP to extract sentences relevant to captioning, treating entity-relevant sentences as positives and unrelated ones as negatives.
\citet{hu2020icecap} fine-tuned VSE++ \citep{faghri2018vse++} for image-text retrieval with a coarse-to-fine attention mechanism for entity identification. 
\citet{zhou2022focus} fine-tuned CLIP for visually grounded entity detection and used open relation extraction for non-visual entities.
\citet{qu2024visually} utilized CLIP for sentence retrieval and developed a Face Naming Module to associate faces with names. 
Building on these ideas, MERGE not only retrieves relevant context but also integrates background knowledge, producing captions that are accurate and journalistically informative.

\paragraph{Methods Incorporating MLLMs}
The advent of GPT-4V~\citep{openai2023gpt4} has significantly advanced MLLMs, achieving impressive results across diverse tasks.
For instance, InstructBLIP~\citep{dai2023instructblip} fuses visual and language features via a Q-Former architecture and improves cross-task generalization through instruction tuning.
LLaVA~\citep{liu2023visual} uses linear projection layers for vision-language alignment and excels on GPT-4-generated multimodal benchmarks.
Advanced models like Qwen2-VL~\citep{Yang2024qwen2}, InternVL2~\citep{chen2024internvl, chen2024far}, and Llama3.2-Vision~\citep{meta2024Llama} surpass GPT-4V, showing versatility in fields like medical diagnosis and autonomous driving.
In news image captioning, \citet{xu2024cross} enhanced LLaVA's performance with confidence-aware prompts and refined CLIP with a matching-score comparative loss. 
EAMA~\citep{zhang2024entity} adapted InstructBLIP for news captioning, introducing alignment tasks to improve multimodal understanding. 
MERGE goes further by integrating retrieval-augmented multimodal knowledge, addressing deeper challenges in entity grounding and contextual completeness.

\section{The MERGE Framework}
\label{sec:methods}

\subsection{Overview}
\label{sec:methods:overview}

MERGE is a multimodal RAG framework developed to advance news image captioning by seamlessly integrating external multimodal data with structured knowledge.
As depicted in Figure \ref{fig:model_overview}, MERGE contains three core components:
\begin{itemize}
  \item \textbf{Entity-centric Multimodal Knowledge Base (EMKB):} 
  Consolidates named entities, images, and background knowledge to bridge information gaps and strengthen contextual grounding.
  
  \item \textbf{Hypothesis Caption-guided Multimodal Alignment (HCMA):} 
  Achieves fine-grained sentence-level alignment between visual and textual inputs through a three-stage Chain-of-Thought (CoT) prompting process.
  
  \item \textbf{Retrieval-driven Multimodal Knowledge Integration (RMKI):} 
  Improves visual-entity grounding by matching visual cues to entities and dynamically constructing background knowledge graphs from EMKB.
\end{itemize}

\subsection{EMKB}
\label{sec:methods:knowledge_base}

EMKB consolidates multimodal data and contextual knowledge as the foundation of MERGE.

\paragraph{Entity Extraction and Image Collection}
We extract entities (e.g., celebrity names, locations, artworks, landmarks, and buildings) from the GoodNews~\citep{biten2019good} and NYTimes800k~\citep{tran2020transform} datasets using spaCy~\citep{honnibal2017spacy}, and expand this set via an LLM.
For each entity, we collect a Wikipedia image and augment it with up to five images from Google Search.
To cover less common or missing celebrity entities, we also incorporate four public datasets (see Appendix B), capping each entity at five images.
This results in a visually rich, entity-centric dataset that supports precise visual-entity alignment.

\paragraph{Background Knowledge Acquisition}
Beyond visual data, EMKB captures deeper contextual knowledge. 
Background information for each entity is extracted from Wikipedia and IMDb and structured into subgraphs using LLMs with a crafted prompt $p_k$ (see Appendix A.1).
Unlike static knowledge graphs~\citep{alberts2021visualsem, wang2019richpedia}, these subgraphs are dynamically retrieved during caption generation, enabling news-specific knowledge integration.

\paragraph{EMKB Formulation}
Formally, the EMKB $\bm{B}$ is defined as:
\begin{equation} \label{eq:dataset}
  \bm{B} = \{(\bm{e}_i, \{\bm{I}_{j}\}, \bm{b}_i, \bm{G}_{sub}^i)\}_{i=1}^N,
\end{equation}
where $\bm{e}_i$ is the $i$-th entity, $\{\bm{I}_{j}\}$ are its associated images, $\bm{b}_i$ denotes background knowledge, and $\bm{G}_{sub}^i$ is its structured knowledge subgraph.
This formulation enhances contextual grounding, aids in entity disambiguation, and facilitates cross-modal alignment. 
Its structure and components are depicted in Figure~\ref{fig:EMKB}.
Implementation details and the updating mechanism for EMKB are provided in Appendix B.

\begin{figure}[t]
  \centering
  \includegraphics[width=0.99\columnwidth]{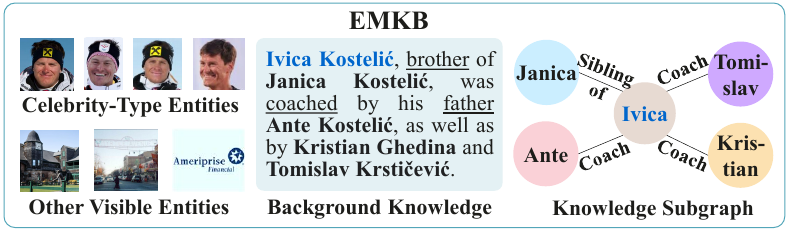}
  \caption{Architecture of EMKB, illustrating named entities, their images, background knowledge, and knowledge subgraphs, which support context-rich news image captioning.}
  \label{fig:EMKB}
\end{figure}

\subsection{HCMA}
\label{sec:methods:alignment}

HCMA tackles sentence-level cross-modal alignment via a three-stage CoT prompting process.

\paragraph{Stage 1: Hypothesis Caption Generation}
A hypothesis caption $\bm{\hat{h}}$ is generated via a crafted prompt $p_h$, encapsulating visual and textual cues from the input image $\bm{I}$ and article $\bm{T}$.
The MLLM first extracts key sentences reflecting the central themes and visual content of $\bm{T}$. 
Let $\hat{h}_{<i}$ be the partial caption up to token $i$.
The model then iteratively generates each word $\hat{h}_i$:
\begin{equation} \label{eq:hypocapgen}
  \hat{h}_i = \text{MLLM}(p_h, \hat{h}_{<i}, \bm{I}, \bm{T}).
\end{equation}

\paragraph{Stage 2: Relevant Sentence Selection}
HCMA refines context by selecting the most relevant sentences $\bm{S}$ from $\bm{T}$ using the hypothesis caption $\bm{\hat{h}}$ and image $\bm{I}$ as anchors. 
The selection uses a dedicated prompt $p_s$:
\begin{equation} \label{eq:sensel}
  \bm{S} = \text{MLLM}(p_s, \bm{\hat{h}}, \bm{I}, \bm{T}),
\end{equation}
where $|\bm{S}| \leq 5$ balances informativeness and efficiency, ensuring alignment with both visual and textual content.

\paragraph{Stage 3: Global Summary Generation}
While $\bm{S}$ and $\bm{\hat{h}}$ capture localized context, they can miss broader connections, as shown in Table \ref{tab:ablation-results}.
To capture a global perspective and manage input length, HCMA generates a concise global summary $\bm{U}$ from $\bm{T}$ using prompt $p_g$:
\begin{equation} \label{eq:summary}
  \bm{U} = \text{MLLM}(p_g, \bm{T}),
\end{equation}
with $\bm{U}$ limited to 100 words for brevity and clarity.

\paragraph{Remarks}
By integrating local context from $\bm{S}$ and global context from $\bm{U}$, HCMA produces captions that are precise and contextually rich.
Details of prompts $p_h$, $p_s$, and $p_g$ are provided in Appendix A.2.

\subsection{RMKI}
\label{sec:methods:retrieval}

At the entity level, RMKI strengthens visual-entity grounding through two Retrieval-Augmented Strategies (RAS) within the EMKB $\bm{B}$: entity matching and background knowledge graph construction. 

\paragraph{RAS 1: Entity Matching}
RMKI matches visual cues in image $\bm{I}$ to entities stored in EMKB $\bm{B}$ via two pathways:
\begin{itemize}
  \item \textbf{Face Images:} 
  Faces detected in $\bm{I}$ are encoded as feature vectors $\bm{F}$ using InsightFace.\footnote{\url{https://github.com/deepinsight/insightface}} 
  For each vector $\bm{y} \in \bm{F}$, RMKI computes cosine similarity against face vectors $\bm{x}_j$ from EMKB images $\bm{I}_j \in \bm{B}$:
  \begin{equation} \label{eq:facerecon}
    j^* = \textstyle \argmax_{j} \cos(\bm{x}_j, \bm{y}).
  \end{equation}
  The matched entities from $\bm{I}_{j^*}$ are then extracted to form the entity set $\bm{E}$.

  \item \textbf{Non-Face Images:} 
  For images without faces, RMKI leverages CLIP's image encoder to generate visual embeddings. 
  Cosine similarity is used to identify the closest matching images $\bm{I}_j \in \bm{B}$, yielding the entity set $\bm{E}$.
\end{itemize}

\begin{algorithm}[t]
  \caption{Background KG Construction}
  \label{alg:background}
  \KwIn{EMKB $\bm{B}$, relevant sentences $\bm{S}$}
  \KwOut{Background knowledge graph $\bm{G}$}
  $\bm{E}_{sen} \leftarrow$ Spacy($\bm{S}$);~$\bm{R} \leftarrow$ LLM($p_r$, $\bm{E}_{sen}$)\;
  $\bm{G}_{base} \leftarrow$ ConstructBaseGraph($\bm{E}_{sen}$, $\bm{R}$)\;
  $\bm{\Phi} \leftarrow \varnothing$\Comment*[r]{Store knowledge subgraphs}
  \ForEach{$\bm{e} \in \bm{E}_{sen}$}{
    $\bm{G}_{sub}^{i} \leftarrow$ Retrieve($\bm{e}$, $\bm{B}$);~$\bm{\Phi} \leftarrow \bm{\Phi} \cup \{\bm{G}_{sub}^{i}\}$\;
  }
  $\bm{G} \leftarrow$ IntegrateGraph($\bm{G}_{base}$, $\bm{\Phi}$)\;
  \Return $\bm{G}$\;
\end{algorithm}
\setlength{\textfloatsep}{1.0em}

\paragraph{RAS 2: Background Knowledge Graph (KG) Construction}
To enrich contextual understanding, RMKI constructs a background KG for entities identified in the relevant sentences $\bm{S}$. 
This process unfolds in four steps:
\begin{itemize}
  \item \textbf{Named Entity Recognition (NER):} Identify named entities $\bm{E}_{sen}$ within $\bm{S}$ using spaCy. 
  
  \item \textbf{Relation Extraction:} Use LLMs with a dedicated prompt $p_r$ to extract relations $\bm{R}$ among entities in $\bm{E}_{sen}$, forming the base relation graph $\bm{G}_{base}$.

  \item \textbf{Subgraph Retrieval:} For each entity $\bm{e} \in \bm{E}_{sen}$, retrieve the knowledge subgraph $\bm{G}_{sub}^{i}$ from the EMKB $\bm{B}$ and aggregate them into a set $\bm{\Phi}$.
  
  \item \textbf{Graph Integration:} Integrate subgraphs $\bm{\Phi}$ into $\bm{G}_{base}$, deduplicating overlapping nodes and edges, to produce the final knowledge graph $\bm{G}$.
\end{itemize}

\begin{table*}[t]
\centering
\small
\renewcommand{\arraystretch}{1.0}
\resizebox{0.99\textwidth}{!}{
\begin{tabular}{cl*{7}{c}}
  \toprule
  &  & \multicolumn{4}{c}{\textbf{Caption Quality}} & \multicolumn{3}{c}{\textbf{Named Entity Accuracy}} \\ \cmidrule(lr){3-6} \cmidrule(lr){7-9} 
  \multirow{-2.5}{*}{\textbf{Dataset}} & \multirow{-2.5}{*}{\textbf{Method}} & \textbf{BLEU-4 $\uparrow$} & \textbf{METEOR $\uparrow$} & \textbf{ROUGE $\uparrow$} & \textbf{CIDEr $\uparrow$} & \textbf{Precision $\uparrow$} & \textbf{Recall $\uparrow$} & \textbf{F1-score $\uparrow$} \\
  \midrule
  
  \multirow{14.5}{*}{\rotatebox{90}{GoodNews}} 
  & \citet{biten2019good}  & 0.89 & 4.37 & 12.20 & 13.10 & 8.23 & 6.06 & 6.98 \\
  & ICECAP~\citep{hu2020icecap}     & 1.96 & 6.01 & 15.70 & 26.08 & - & - & 12.03 \\
  & Tell~\citep{tran2020transform}  & 6.05 & 10.30 & 21.40 & 53.80 & 22.20 & 18.70 & 20.30 \\ 
  & JoGANIC~\citep{yang2021journalistic} & 6.83 & 11.25 & 23.05 & 61.22 & 26.87 & 22.05 & 24.22 \\ 
  & \citet{liu2021visual}  & 6.10 & 8.30 & 21.60 & 55.40 & 22.90 & 19.30 & 20.95 \\
  & \citet{zhou2022focus}  & 6.30 & - & 22.40 & 60.30 & 24.20 & 20.90 & 22.43 \\
  & NewsMEP~\citep{zhang2022fine}   & 8.30 & 12.23 & 23.17 & 63.99 & 23.43 & 23.24 & 23.33 \\ 
  & \citet{kalarani2023let}   & 7.14 & 11.21 & 24.30 & 72.33 & 24.37 & 20.09 & 22.02 \\  
  & \citet{zhao2024boosting}        & 8.31 & 12.32 & 23.22 & 64.15 & - & - & 23.39 \\
  & \citet{xu2024rule}     & 8.18 & 12.50 & 23.56 & 71.58 & 25.51 & 23.68 & 24.56 \\  
  & \citet{qu2024visually} & 8.60 & 12.39 & 23.38 & 71.96 & 24.30 & 25.54 & 24.90 \\  
  & \citet{xu2024cross}    & 8.49 & 12.88  & 26.22 & 83.52 & \underline{30.19} & 26.57 & \underline{28.26} \\ 
  & EAMA \citep{zhang2024entity}    & \underline{10.04} & \underline{13.95} & \underline{27.06} & \underline{87.70} & 27.58 & \underline{28.92} & 28.23 \\
  \cmidrule{2-9}
  & MERGE & \textbf{10.19}  & \textbf{14.31} & \textbf{28.02} & \textbf{94.54} & \textbf{32.89} & \textbf{31.93} & \textbf{32.40} \\
  \midrule
  
  \multirow{12.5}{*}{\rotatebox{90}{NYTimes800k}}
  & Tell \citep{tran2020transform}  & 6.30 & 10.30 & 21.70 & 54.40 & 24.60 & 22.20 & 23.34 \\
  & JoGANIC \citep{yang2021journalistic}  & 6.79 & 10.93 & 22.80 & 59.42 & 28.63 & 24.49 & 26.40 \\ 
  & \citet{liu2021visual}    & 6.40 & 8.10 & 21.90 & 56.10 & 24.80 & 22.30 & 23.48 \\ 
  & \citet{zhou2022focus}  & 7.00 & - & 22.90 & 63.60 & 29.80 & 25.90 & 27.71 \\ 
  & NewsMEP \citep{zhang2022fine}   & 9.57 & 13.02 & 23.62 & 65.85 & 26.61 & 28.57 & 27.56 \\ 
  & \citet{kalarani2023let}   & 7.54 & 11.27 & 23.28 & 66.41 & 28.11 & 23.25 & 25.45 \\ 
  & \citet{zhao2024boosting} & 9.53 & 13.30 & 23.89 & 66.43 & - & - & 27.71 \\ 
  & \citet{xu2024rule}       & 9.41 & 13.10 & 24.42 & 72.29 & 28.15 & 28.80 & 28.47 \\ 
  & \citet{qu2024visually}   & 9.24 & 12.57 & 23.44 & 71.65 & 26.88 & 28.59 & 27.71 \\ 
  & \citet{xu2024cross}      & 9.07 & 13.17  & 26.48 & 83.72 & \textbf{32.38} & 30.08 & \underline{31.19} \\ 
  & EAMA \citep{zhang2024entity} & \underline{11.03} & \underline{14.22} & \underline{27.15} & \underline{87.00} & 29.79 & \underline{32.24} & 30.97 \\ 
  \cmidrule{2-9}
  & MERGE & \textbf{11.47}  & \textbf{14.94} & \textbf{27.51} & \textbf{88.16} & \underline{31.87} & \textbf{36.04} & \textbf{33.83} \\
  \midrule

  \multirow{5.5}{*}{\rotatebox{90}{Visual News}}
  & Tell~\citep{tran2020transform}  & 9.60 & - & 22.80 & 83.80 & 23.70 & 19.20 & 21.21 \\ 
  & Visual News Captioner~\citep{liu2021visual}  & 5.30 & 8.20 & 17.90 & 50.50 & 19.70 & 17.60 & 18.59 \\ 
  & \citet{zhou2022focus}   & \underline{11.60} & - & \underline{25.00} & \underline{107.60} & \underline{26.20} & \underline{21.20} & \underline{23.44} \\ 
  & \citet{kalarani2023let} & 6.91 & \underline{10.54} & 21.29 & 65.14 & 19.31 & 19.90 & 19.60 \\ 
  \cmidrule{2-9}
  & MERGE & \textbf{14.77}  & \textbf{15.72} & \textbf{28.26} & \textbf{127.77} & \textbf{29.88} & \textbf{29.45} & \textbf{29.66} \\
  \bottomrule
\end{tabular}}
\caption{Experimental results on GoodNews, NYTimes800k, and Visual News, comparing MERGE with baseline methods.}
\label{tab:main-results}
\end{table*}
\setlength{\textfloatsep}{1.0em}

This dynamic retrieval and graph integration enables MERGE to enrich generated captions with precision, entity-specific knowledge, improving both contextual coherence and factual accuracy.
Algorithm \ref{alg:background} details this process, with prompt $p_r$ provided in Appendix A.3.

\subsection{Caption Generation}
\label{sec:methods:generation}

Given an image $\bm{I}$ and a news article $\bm{T}$, MERGE proceeds as follows:
First, HCMA generates a hypothesis caption $\bm{\hat{h}}$, selects relevant sentences $\bm{S}$, and creates a global summary $\bm{U}$.
Second, RMKI matches $\bm{I}$ to entities $\bm{E}$ and builds a background knowledge graph $\bm{G}$.
Finally, InstructBLIP~\citep{dai2023instructblip} is employed for caption generation, enhanced with a 4-layer Graph Attention Network (GAT)~\citep{velickovic2018graph} to encode $\bm{G}$, integrating multimodal inputs: $\bm{X} = \{\bm{I}, \bm{\hat{h}}, \bm{S}, \bm{U}, \bm{E}, \bm{G}\}$ to produce the final caption $\hat{\bm{c}}$:
\begin{equation}\label{eq:caption_output}
  \hat{\bm{c}} = \text{MLLM}(\bm{X}; \bm{\theta}),
\end{equation}
where $\bm{\theta}$ denotes the parameters of MLLM. 
The model is trained to minimize the Cross-Entropy (CE) loss, which measures the negative log-likelihood of the ground truth caption $\bm{c}$ conditioned on the multimodal inputs:
\begin{equation} \label{eq:loss_ce}
  \mathcal{L}_{CE} = \textstyle -\sum_{i=1}^{|\bm{c}|} \log P(c_i \mid c_{<i}, \bm{X}). 
\end{equation}

MERGE combines textual, visual, and structured knowledge, resulting in comprehensive, contextually aligned news image captions, as validated in Section \ref{sec:expt:main_results}.

\section{Experiments}
\label{sec:expt}

\begin{table*}[t]
\centering
\small
\renewcommand{\arraystretch}{1.0}
\resizebox{0.99\textwidth}{!}{
\begin{tabular}{cl*{7}{c}}
  \toprule
  &  & \multicolumn{4}{c}{\textbf{Caption Quality}} & \multicolumn{3}{c}{\textbf{Named Entity Accuracy}} \\ \cmidrule(lr){3-6} \cmidrule(lr){7-9} 
  \multirow{-2.5}{*}{\textbf{Dataset}} & \multirow{-2.5}{*}{\textbf{Method}} & \textbf{BLEU-4 $\uparrow$} & \textbf{METEOR $\uparrow$} & \textbf{ROUGE $\uparrow$} & \textbf{CIDEr $\uparrow$} & \textbf{Precision $\uparrow$} & \textbf{Recall $\uparrow$} & \textbf{F1-score $\uparrow$} \\
  \midrule
  
  \multirow{11}{*}{\rotatebox{90}{GoodNews}} 
  & InstructBLIP (w/o FT) & 3.46 & 8.41 & 13.82 & 24.42 & 16.28 & 14.21 & 15.17 \\
  \cmidrule{2-9}
  & InstructBLIP (w/ FT) & 8.57 & 12.64 & 24.53 & 84.8 & 32.23 & 27.64 & 29.76 \\ [0.5ex]
  & \quad + HCMA (w/ Stage 1) & 9.33 & 13.50 & 26.31 & 84.83 & 30.10 & 29.16 & 29.62 \\
  & \quad + HCMA (w/ Stage 1+2) & 9.49 & 13.33 & \underline{26.63} & 85.24 & 30.73 & 29.19 & 29.94 \\
  & \quad + HCMA (w/ Stage 1+2+3) & 9.32 & 12.53 & 26.59 & 86.08 & 30.77 & 29.30 & 30.02 \\
  & \quad + RMKI \& EMKB (w/ RAS 1) & 9.62 & 13.74 & 25.75 & \underline{91.52} & \textbf{33.72} & 30.98 & \underline{32.29} \\
  & \quad + RMKI \& EMKB (w/ RAS 2) & 9.20 & 13.12 & 26.52 & 85.79 & 28.91 & 30.94 & 29.89 \\
  & \quad + RMKI \& EMKB (w/ RAS 1+2)& \underline{9.73} & \underline{13.79} & 25.76 & 91.36 & \underline{33.37} & \underline{31.27} & \underline{32.29} \\[0.5ex]
  \cmidrule{2-9}
  & MERGE & \textbf{10.19} & \textbf{14.31} & \textbf{28.02} & \textbf{94.54} & 32.89 & \textbf{31.93} & \textbf{32.40} \\
  \midrule
  
  \multirow{11}{*}{\rotatebox{90}{NYTimes800k}} 
  & InstructBLIP (w/o FT) & 3.77 & 8.14 & 13.16 & 22.33 & 13.66 & 15.68 & 14.60  \\
  \cmidrule{2-9}
  & InstructBLIP (w/ FT) & 9.16 & 12.68 & 24.70 & 73.68 & 28.72 & 29.69 & 29.20 \\ [0.5ex]
  & \quad + HCMA (w/ Stage 1) & 9.15 & 13.07 & 25.04 & 75.40 & 30.72 & 30.92 & 30.82 \\
  & \quad + HCMA (w/ Stage 1+2) & 10.27 & 14 15 & 25.51 & 76.81 & 27.74 & 32.91 & 30.10 \\
  & \quad + HCMA (w/ Stage 1+2+3) & 10.01 & 13.62 & 25.79 & 80.14 & 30.23 & 31.83 & 31.01 \\
  & \quad + RMKI \& EMKB (w/ RAS 1) & \underline{11.09} & \underline{14.48} & 26.53 & 81.04 & 30.53 & \underline{34.82} & 32.53 \\
  & \quad + RMKI \& EMKB (w/ RAS 2) & 10.39 & 14.07 & 25.16 & 74.43 & 27.80 & 32.59 & 30.01 \\
  & \quad + RMKI \& EMKB (w/ RAS 1+2) & 10.75 & 14.17 & \underline{26.61} & \underline{82.99} & \underline{31.76} & 34.28 & \underline{32.97} \\[0.5ex]
  \cmidrule{2-9}
  & MERGE & \textbf{11.47} & \textbf{14.94} & \textbf{27.51} & \textbf{88.16} & \textbf{31.87} & \textbf{36.04} & \textbf{33.83} \\
  \midrule

  \multirow{11}{*}{\rotatebox{90}{Visual News}} 
  & InstructBLIP (w/o FT) & 1.67  & 3.85  & 5.92  & 7.63  & 5.57  & 6.29  & 5.91  \\
  \cmidrule{2-9}
  & InstructBLIP (w/ FT) & 14.00 & 15.10  & 27.47  & 103.48  & 26.20  & 28.04 & 27.09  \\ [0.5ex]
  & \quad + HCMA (w/ Stage 1) & 13.39  & 15.12  & 25.60  & 103.95  & 25.81  & 29.33  & 27.46  \\
  & \quad + HCMA (w/ Stage 1+2) & 14.37  & 15.34  & 27.33  & 105.28  & 28.77  & 26.61  & 27.65  \\
  & \quad + HCMA (w/ Stage 1+2+3) & 14.57  & 15.44  & 27.53  & 108.97  & 27.32  & \textbf{29.56}  & \underline{28.40}  \\
  & \quad + RMKI \& EMKB (w/ RAS 1) & 13.33  & 14.90  & 27.37  & 116.84  & \underline{29.74}  & 25.17 & 27.26  \\
  & \quad + RMKI \& EMKB (w/ RAS 2) & \textbf{14.95}  & \underline{15.46}  & \underline{27.83}  & 113.41  & 27.06  & 28.99  & 27.99  \\
  & \quad + RMKI \& EMKB (w/ RAS 1+2) & 14.12  & 15.03  & 27.30  & \underline{123.50}  & 26.36  & 28.01  & 27.11  \\ [0.5ex]
  \cmidrule{2-9}
  & MERGE & \underline{14.77}  & \textbf{15.72} & \textbf{28.26} & \textbf{127.77} & \textbf{29.88} & \underline{29.45} & \textbf{29.66} \\
  \bottomrule
\end{tabular}}
\caption{Ablation results on GoodNews, NYTimes800k, and Visual News, highlighting the impact of MERGE's components.}
\label{tab:ablation-results}
\end{table*}

\begin{figure*}[ht]
  \centering
  \includegraphics[width=0.99\textwidth]{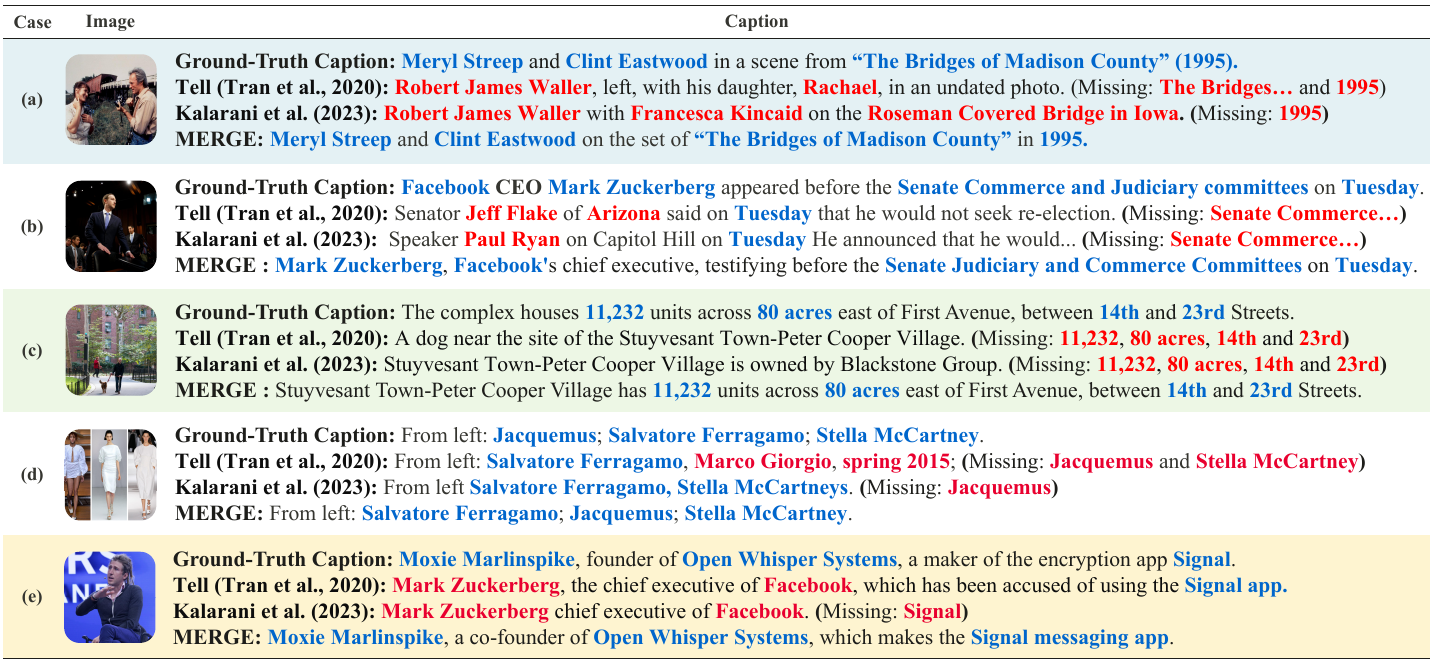}
  \caption{Case study on GoodNews. Entities correctly identified by MERGE are depicted in blue, while errors are shown in red.}
  \label{fig:case_study_goodnews}
\end{figure*}

\subsection{Experiment Setup}
\label{sec:expt:setup}

\paragraph{Datasets}
We assess MERGE on GoodNews \citep{biten2019good}, NYTimes800k \citep{tran2020transform}, and Visual News \citep{liu2021visual}.
Visual News is excluded from EMKB construction and used to evaluate MERGE's generalization.
Additional details are in Appendix C.

\paragraph{Metrics}
We assess caption quality using BLEU-4~\citep{papineni2002bleu}, METEOR~\citep{denkowski2014meteor}, ROUGE~\citep{lin2004rouge}, and CIDEr~\citep{vedantam2015cider}.
Additionally, named entity accuracy is measured via Precision, Recall, and F1-score using spaCy~\citep{honnibal2017spacy}.

\paragraph{Baselines}
We compare MERGE (implementation details in Appendix D) with state-of-the-art baselines below.
\begin{itemize}
  \item \textbf{Methods Processing Original Articles:} \citet{biten2019good}, Tell~\citep{tran2020transform}, JoGANIC~\citep{yang2021journalistic}, NewsMEP~\citep{zhang2022fine}, alongside recent approaches like \citet{kalarani2023let}, \citet{xu2024rule} and \citet{zhao2024boosting}. 

  \item \textbf{Methods Using Extracted Relevant Contexts:} ICECAP~\citep{hu2020icecap}, \citet{zhou2022focus}, and \citet{qu2024visually}. 

  \item \textbf{Methods Incorporating MLLMs:} \citet{xu2024cross} and EAMA~\citep{zhang2024entity}.
\end{itemize}

\subsection{Comparison Results of Different Baselines}
\label{sec:expt:main_results}

Table \ref{tab:main-results} summarizes how MERGE performs against baseline methods on GoodNews, NYTimes800k, and Visual News.

\paragraph{Caption Quality}
MERGE achieves state-of-the-art results across nearly all metrics.
Compared to the strongest baseline EAMA, MERGE improves CIDEr by +6.84 on GoodNews and +1.16 on NYTimes800k, while remaining competitive on other metrics.
The smaller margin on NYTimes800k reflects its higher complexity, with articles being nearly twice as long as GoodNews, hindering relevant content extraction.

These gains highlight prior methods' limitations in cross-modal alignment and visual-entity grounding.
Baselines like \citet{qu2024visually} and \citet{zhang2024entity} rely on sentence retrieval but focus on either image descriptions or entity-rich text, causing suboptimal alignment.
Moreover, existing methods often fail to robustly link entities across modalities, degrading caption quality.

In contrast, MERGE's HCMA module effectively selects relevant context, while RMKI and EMKB integrate multimodal RAG to balance visual and textual information, achieving more precise entity alignment. These advantages are further demonstrated in Section \ref{sec:expt:case}.

\paragraph{Named Entity Accuracy}
MERGE also sets new benchmarks for NER.
On GoodNews, MERGE surpasses \citet{xu2024cross} by +2.70 precision and +4.14 F1-score, and exceeds EAMA by +3.01 recall, validating EMKB and RMKI effectiveness.
On NYTimes800k, MERGE achieves +2.64 F1-score improvement, though precision slightly trails \citet{xu2024cross} due to their additional training data (20\% training set plus full validation) and knowledge distillation for fine-tuning.
In contrast, MERGE's HCMA requires no extra training, ensuring adaptability.

\paragraph{Generalization Test}
Table~\ref{tab:main-results} also shows MERGE's generalization on Visual News.
MERGE outperforms the second-best method, \citet{zhou2022focus}, by +20.17 in CIDEr and +6.22 in F1-score. 
This significant improvement underscores how EMKB provides broad coverage for news image captioning, even for datasets not included in its construction.

\subsection{Ablation Study}
\label{sec:expt:ablation}

Table \ref{tab:ablation-results} reports the ablation results, quantifying the contributions of MERGE’s core components.

\paragraph{Impact of MLLMs}
MERGE adopts InstructBLIP as its MLLM backbone (Section \ref{sec:methods:generation}).
\textbf{InstructBLIP (w/o FT)} shows general-purpose MLLMs fall short on domain-specific news captioning when used zero-shot.
After task-specific tuning, \textbf{InstructBLIP (w/ FT)} achieves substantial gains over conventional baselines (Table~\ref{tab:main-results}), validating fine-tuning effectiveness.
We further evaluate other advanced MLLMs and compare efficiency in Appendices F--H, confirming MERGE's flexibility and efficiency.

\paragraph{Impact of HCMA}
Introducing the HCMA component incrementally improves both generation quality and entity recognition.
\textbf{HCMA (w/ Stage 1)} introduces hypothesis captioning, offering notable gains in both CIDEr and F1-scores.
Incorporating sentence selection in \textbf{HCMA (w/ Stage 1+2)} further enhances alignment by focusing on semantically relevant context.
Finally, including global summary generation in \textbf{HCMA (w/ Stage 1+2+3)} achieves the best overall performance, highlighting the importance of progressively structured textual grounding.

\paragraph{Impacts of RMKI and EMKB}
The RMKI and EMKB jointly drive significant improvements.
\textbf{RMKI \& EMKB (w/ RAS 1)}, which supports visual-entity alignment through entity matching, enhances precision in NER.
\textbf{RMKI \& EMKB (w/ RAS 2)} leverages background knowledge graphs, improving contextual grounding and recall.
Combining both in \textbf{RMKI \& EMKB (w/ RAS 1+2)} leads to the highest F1 and CIDEr scores, confirming the complementary benefits of structured retrieval and visual grounding.

\paragraph{Overall Impact}
These findings confirm that MERGE's performance stems from the synergy of its key components: a fine-tuned MLLM backbone, multistage cross-modal alignment, and dual-stream retrieval from structured and unstructured knowledge. 
Together, they enable MERGE to effectively tackle the core challenges of news image captioning.

\subsection{Case Study}
\label{sec:expt:case}

Figure~\ref{fig:case_study_goodnews} showcases MERGE's outputs on GoodNews examples, highlighting three key capabilities:
\begin{itemize}
  \item \textbf{Information Enhancement (via EMKB):} 
  In case (a), MERGE correctly identifies \texttt{Clint Eastwood} using EMKB, although his name is absent from the article.
  
  \item \textbf{Fine-grained Cross-modal Alignment (via HCMA):} 
  Cases (b) and (c) demonstrate accurate alignment between visual and textual cues, successfully incorporating details like \texttt{Senate Commerce and Judiciary committees}, \texttt{11,232 units}, and \texttt{80 acres}. 
  
  \item \textbf{Precise Visual-Entity Alignment (via RMKI):}
  Cases (d) and (e) demonstrate how RMKI enables MERGE to distinguish among multiple individuals and visually similar subjects, ensuring correct entity grounding.
\end{itemize}

These results highlight MERGE's ability to enrich missing context, align cross-modal details, and resolve complex visual references, which are crucial for real-world news captioning. Additional examples from NYTimes800k and Visual News are provided in Appendix I.

\section{Conclusions}
\label{sec:conclusions}

In this work, we introduced MERGE, a novel multimodal entity-aware RAG framework for news image captioning.
By building an entity-centric multimodal knowledge base and integrating structured, visual, and textual information, MERGE effectively tackles key challenges in contextual grounding, cross-modal alignment, and visual-entity association.
Extensive experiments on GoodNews, NYTimes800k, and Visual News show that MERGE consistently surpasses strong baselines, achieving state-of-the-art results in both caption generation and named entity recognition.
These findings underscore the power of multimodal RAG for complex, knowledge-intensive vision-language tasks and mark an important step forward in bridging multimodal understanding and real-world journalistic applications.

\section*{Acknowledgements}
We sincerely thank the anonymous reviewers for their insightful and constructive feedback, which greatly improved this paper.
This work was supported by the National Natural Science Foundation of China (NSFC) under Grant Nos. 62125201 and U24B20174.

\bibliography{aaai2026}
\appendix

\section{Prompts}
\label{sec:prompts}

We detail the prompts that power MERGE's retrieval-augmented processes, covering knowledge subgraph construction in EMKB, three-stage CoT prompting in HCMA, and relation extraction in RMKI.

\subsection{Knowledge Subgraph Construction Prompt}
\label{sec:prompts:knowledge}

\begin{figure}[t]
  \centering
  \includegraphics[width=0.99\columnwidth]{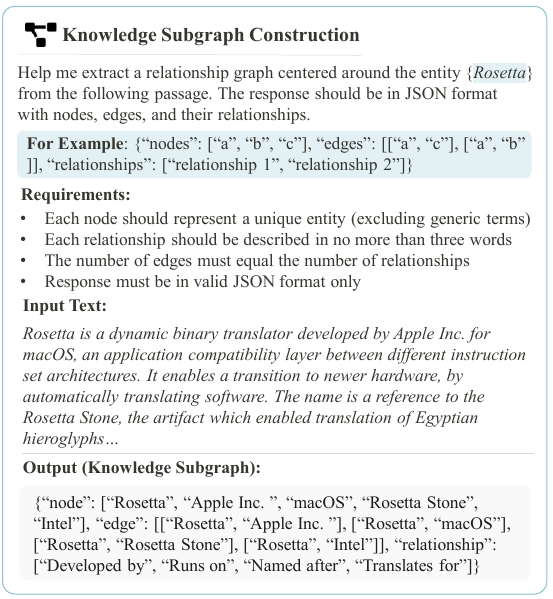}
  \caption{Prompt template ($p_k$) used in EMKB to construct knowledge subgraphs.}
  \label{fig:prompt_subgraph}
\end{figure}

Figure \ref{fig:prompt_subgraph} illustrates the Knowledge Subgraph Construction prompt $p_k$, which extracts and structures entity-centric knowledge to enhance contextual grounding in news image captioning.
Each subgraph is centered around a specific entity and built from background knowledge stored in EMKB, ensuring cohesive and semantically rich representations of entity relationships.

\paragraph{Prompt Design}
To ensure subgraphs are informative and precise, we impose three key constraints:
\begin{itemize}
  \item \textbf{Uniqueness:} Each node in the subgraph represents a distinct entity, preventing redundancy.
  
  \item \textbf{Controlled Relationship Length:} Relationships are kept concise and restricted in length to reduce noise and enhance semantic clarity.
  
  \item \textbf{Semantic Specificity:} Generic terms (e.g., \texttt{system}, \texttt{method}, and \texttt{application}) are excluded to preserve contextual relevance.
\end{itemize}

The subgraphs are structured in a standardized JSON format, containing three main components:
\begin{itemize}
  \item \textbf{Nodes:} Entities extracted from background text.   
  \item \textbf{Edges:} Directed links indicating relationships between entities.
  \item \textbf{Relationships:} Descriptions of how entities are related, ensuring explicit semantic connections.
\end{itemize}

\paragraph{Example Case}
Figure~\ref{fig:prompt_subgraph} provides an example where the entity \texttt{Rosetta} connects to related concepts like \texttt{Apple Inc}, \texttt{macOS}, and \texttt{Intel}, effectively capturing technical and historical context for precise retrieval and integration in MERGE.

\paragraph{Remarks}
By structuring entity relationships and standardizing knowledge retrieval, $p_k$ significantly enhances contextual grounding and supports robust cross-modal understanding in news image captioning.

\subsection{Three-Stage CoT Prompt}
\label{sec:prompts:cot}

\begin{figure}[t]
  \centering
  \includegraphics[width=0.99\columnwidth]{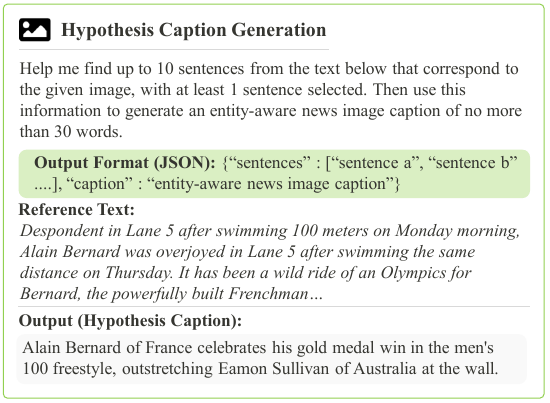}
  \caption{Stage 1 prompt template $p_h$ in HCMA for hypothesis caption generation.}
  \label{fig:prompt_HCMA_stage_1}
\end{figure}

\begin{figure}[t]
  \centering
  \includegraphics[width=0.99\columnwidth]{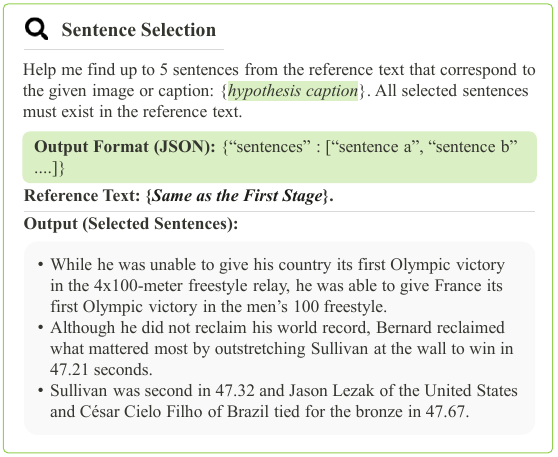}
  \caption{Stage 2 prompt template $p_s$ in HCMA for relevant sentence selection.}
  \label{fig:prompt_HCMA_stage_2}
\end{figure}

\begin{figure}[t]
  \centering
  \includegraphics[width=0.99\columnwidth]{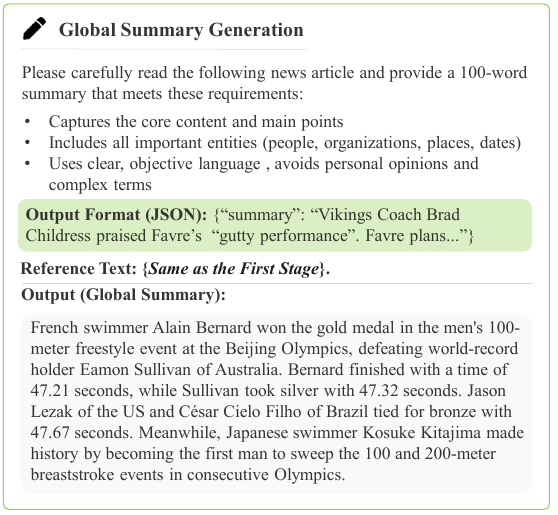}
  \caption{Stage 3 prompt template $p_g$ in HCMA for global summary generation.}
  \label{fig:prompt_HCMA_stage_3}
\end{figure}

Figures~\ref{fig:prompt_HCMA_stage_1}--\ref{fig:prompt_HCMA_stage_3} present MERGE's three-stage CoT prompting framework, which iteratively refines cross-modal alignment.

\paragraph{Prompt Design}
This framework operates in three stages, each guided by a dedicated prompt:
\begin{itemize}
  \item \textbf{Stage 1: Hypothesis Caption Generation ($p_h$).}
  Generates a concise, entity-aware caption ($\leq$~30 words) by extracting up to 10 key sentences from the news article. 
  The output, in JSON format, serves as the textual foundation for alignment.

  \item \textbf{Stage 2: Sentence Selection ($p_s$).}
  Identifies up to five sentences that bridge the semantic gap between the hypothesis caption and visual content. Outputs are formatted in JSON to streamline integration.

  \item \textbf{Stage 3: Global Summary Generation ($p_r$).}
  Synthesizes a global summary to unify selected sentences and enrich the overall context, thereby enhancing the coherence of the captioning process.
\end{itemize}

\paragraph{Example Case}
Figures~\ref{fig:prompt_HCMA_stage_1}--\ref{fig:prompt_HCMA_stage_3} illustrate the three-stage process using an article about the 2008 Beijing Olympics swimming competition.
In Stage 1, $p_h$ extracts key details of French swimmer Alain Bernard's gold medal victory in the 100m freestyle. 
In Stage 2, $p_s$ chooses three sentences covering Bernard's race progression, specific performance details (e.g., his 47.21-second finish), and the rankings of other competitors.
Finally, in Stage 3, $p_r$ produces a coherent summary that integrates these details with additional contextual background.
Notably, MLLMs sometimes produce JSON formatting errors, particularly when handling special characters like colons. When such parsing issues occur, we re-prompt the MLLM to regenerate the output.

\paragraph{Remarks}
This CoT framework systematically extracts, refines, and integrates information, bridging textual and visual modalities. 
It improves entity awareness, contextual coherence, and cross-modal alignment in MERGE.

\subsection{Relation Extraction Prompt}
\label{sec:prompts:relation}

\begin{figure}[t]
  \centering
  \includegraphics[width=0.99\columnwidth]{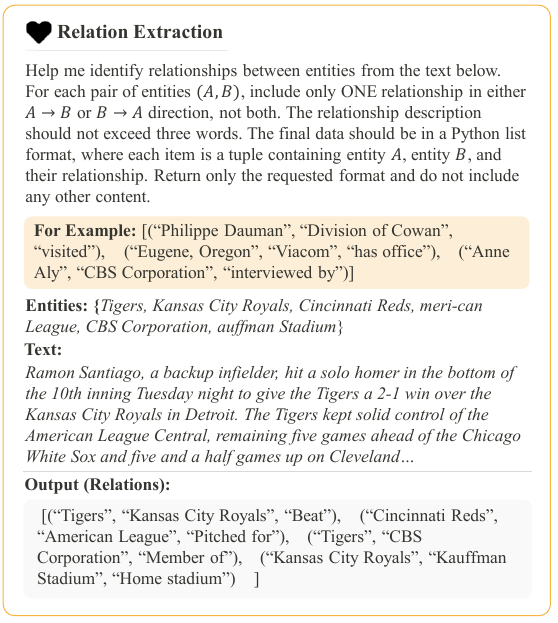}
  \caption{Prompt template $p_r$ used in RMKI for relation extraction.}
  \label{fig:prompt_RMKI}
\end{figure}

Figure \ref{fig:prompt_RMKI} shows the Relation Extraction prompt $p_r$, which identifies entity relationships in news text to construct dynamic background knowledge graphs.

\paragraph{Prompt Design}
The prompt operates under the specific constraints below:
\begin{itemize}
  \item \textbf{Directional Relationships:} 
  Each entity pair $(A, B)$ is assigned to a single directed relationship to avoid redundant bidirectional edges.
  
  \item \textbf{Concise Descriptions:}
  Relationship descriptions are limited to three words for clarity and consistency with the knowledge subgraph format.

  \item \textbf{Structured Entity Extraction:} 
  Named entities are extracted using SpaCy and formatted as comma-separated strings.
  
  \item \textbf{Structured Output:}
  Relationships are returned as Python lists of (source, target, relation) tuples for seamless graph integration.
\end{itemize}

\paragraph{Example Case}
Figure \ref{fig:prompt_RMKI} demonstrates $p_r$ on an article about a baseball game where \texttt{Ramon Santiago's} 10th-inning home run leads the \texttt{Tigers} to defeat the \texttt{Royals} 2-1. 
The prompt successfully extracts key entity relationships, encoding event interactions as structured tuples.

\paragraph{Remarks}
By transforming unstructured news text into structured entity relationships, $p_r$ enables RMKI to construct dynamic background knowledge graphs, thereby deepening contextual understanding in news image captioning.

\begin{table}[t]
\centering
\small
\renewcommand{\arraystretch}{1.1}
\begin{tabular}{lrr}
  \toprule
  \textbf{Dataset} & \textbf{\# Entities} & \textbf{\# Images}  \\ 
  \midrule
  Celebrity-Type Entities$^*$ & 334,806 & 1,589,087       \\
  Other Visible Entities$^*$  & 83,616  & 471,824         \\  
  IMDb-WIKI                   & 60,107  & 84,472          \\
  VGGFace2                    & 8,393   & 36,705          \\
  CACD                        & 1,711   & 4,006           \\
  IMDb-Face                   & 452     & 463             \\
  \midrule
  Total                       & 489,085 & 2,186,557       \\
  \bottomrule
\end{tabular}
\caption{Statistical distribution of entities and their associated number of images in EMKB. Here, $^*$ indicates entities extracted from the GoodNews and NYTimes800k datasets. Other visually representable entities include locations, landmarks, buildings, organizations, artworks, and products.}
\label{table:entity_database}
\end{table}

\begin{table}[t]
\centering
\small
\renewcommand{\arraystretch}{1.1}
\begin{tabular}{llr}
  \toprule
  \textbf{Description} & \textbf{Symbol} & \textbf{Setting} \\
  \midrule
  Maximum context length & $n_{ctx}$ & 1,024 \\
  Maximum output length & $n_{out}$ & 50 \\
  Maximum sequence length & $n_{seq}$ & 4,096 \\
  LoRA rank & $r$ & 16 \\
  LoRA scaling factor & $\alpha$ & 16 \\
  Total batch size & $B$ & 16 \\
  Learning rate (GoodNews) & $\eta_1$ & $3.0 \times 10^{-5}$ \\
  Learning rate (NYTimes800k) & $\eta_2$ & $2.0 \times 10^{-5}$ \\
  Learning rate (Visual News) & $\eta_3$ & $2.0 \times 10^{-5}$ \\
  Epsilon constant & $\epsilon$ & $1.0 \times 10^{-4}$ \\
  Random seed & $s$ & 42 \\
  \bottomrule
\end{tabular}
\caption{Hyperparameters used in MERGE.}
\label{table:hyperparameters}
\end{table}

\begin{table*}[t]
\centering
\small
\renewcommand{\arraystretch}{1.2}
\setlength{\tabcolsep}{3pt}
\resizebox{0.99\textwidth}{!}{%
\begin{tabular}{clccccccccc} 
  \toprule
  \multirow{2.5}{*}{\textbf{Dataset}} & \multirow{2.5}{*}{\textbf{Method}} & \multicolumn{3}{c}{\textbf{PERSON}} & \multicolumn{3}{c}{\textbf{GPE}} & \multicolumn{3}{c}{\textbf{ORG}} \\
  \cmidrule(l){3-5} \cmidrule(l){6-8} \cmidrule(l){9-11}
  & & \textbf{Precision $\uparrow$} & \textbf{Recall $\uparrow$} & \textbf{F1-score $\uparrow$} & \textbf{Precision $\uparrow$} & \textbf{Recall $\uparrow$} & \textbf{F1-score $\uparrow$} & \textbf{Precision $\uparrow$} & \textbf{Recall $\uparrow$} & \textbf{F1-score $\uparrow$} \\
  \midrule
  
  \multirow{5.5}{*}{\rotatebox{90}{GoodNews}}
  & Tell~\citep{tran2020transform} & 29.20 & 23.10  & 25.79 & 25.60 & 24.70 & 25.14 & \underline{22.40} & 18.70 & 20.38 \\
  & NewsMEP~\citep{zhang2022fine} & 32.50 & 29.06  & 30.68 & \underline{26.18} & \underline{30.37} & \underline{28.12} & 22.10 & \underline{22.47} & \underline{22.28} \\
  & \citet{qu2024visually} & 31.00 & 27.42  & 29.10 & 25.99 & 25.62 & 25.80 & 22.21 & 21.51 & 21.85 \\
  & \citet{xu2024rule} & \underline{33.89} & \underline{30.09}  & \underline{31.88} & - & - & - & - & - & - \\
  \cmidrule{2-11}
  & MERGE & \textbf{48.16} & \textbf{42.67} & \textbf{45.25} & \textbf{31.01} & \textbf{35.61} & \textbf{33.15} & \textbf{28.13} & \textbf{28.40} & \textbf{28.26} \\
  \cmidrule{1-11}
  
  \multirow{5.5}{*}{\rotatebox{90}{NYTimes800k}}
  & Tell~\citep{tran2020transform} & 37.30 & 31.10  & 33.92 & 26.80 & 30.60 & 28.57 & 21.90 & 17.20 & 19.26 \\
  & NewsMEP~\citep{zhang2022fine} & 41.32 & 38.51  & 39.87 & 28.24 & \underline{38.76} & \underline{32.67} & \underline{23.10} & \underline{23.51} & \underline{23.30} \\
  & \citet{qu2024visually} & 38.59 & 35.43  & 36.94 & \underline{28.44} & 29.03 & 28.73 & 23.02 & 20.86 & 21.89 \\
  & \citet{xu2024rule} & \underline{41.46} & \underline{39.88} & \underline{40.65} & - & - & - & - & - & - \\
  \cmidrule{2-11}
  & MERGE & \textbf{47.09}  & \textbf{47.51} & \textbf{47.30} & \textbf{31.49} & \textbf{43.62} & \textbf{36.58} & \textbf{26.60} & \textbf{29.24} & \textbf{27.86} \\
  \cmidrule{1-11}

  \multirow{5}{*}{\rotatebox{90}{Visual News}}
  & Tell~\citep{tran2020transform} & 26.85  & 21.67  & 23.98 & \underline{20.12}  & 19.50  & 19.81 & \underline{16.47}  & 12.61  & 14.28  \\
  & Visual News Captioner~\citep{liu2021visual}  & - & - & - & - & - & - & - & - & - \\ 
  & \citet{zhou2022focus}   & - & - & - & - & - & - & - & - & - \\ 
  & \citet{kalarani2023let} & \underline{32.33}  & \underline{27.66}  & \underline{29.81} & 19.41  & \underline{23.01}  & \underline{21.06} & 14.86  & \underline{16.32}  & \underline{15.56}  \\
  \cmidrule{2-11}
  & MERGE & \textbf{39.55}  & \textbf{39.40} & \textbf{39.47} & \textbf{25.30} & \textbf{31.79} & \textbf{28.18} & \textbf{25.89} & \textbf{22.18} & \textbf{23.89} \\
  \bottomrule
\end{tabular}}
\caption{Detailed NER results categorized by three entity types: \textbf{PERSON} (people), \textbf{GPE} (geopolitical entities: countries, cities, and states), and \textbf{ORG} (organizations: companies, agencies, and institutions).}
\label{tab:entity-results}
\end{table*}

\begin{table*}[ht]
\centering
\small
\renewcommand{\arraystretch}{1.2}
\resizebox{0.99\textwidth}{!}{
\begin{tabular}{cl*{7}{c}} 
  \toprule
  &  & \multicolumn{4}{c}{\textbf{Caption Quality}} & \multicolumn{3}{c}{\textbf{Named Entity Accuracy}} \\ \cmidrule(lr){3-6} \cmidrule(lr){7-9} 
  \multirow{-2.5}{*}{\textbf{Dataset}} & \multirow{-2.5}{*}{\textbf{Method}} & \textbf{BLEU-4 $\uparrow$} & \textbf{METEOR $\uparrow$} & \textbf{ROUGE $\uparrow$} & \textbf{CIDEr $\uparrow$} & \textbf{Precision $\uparrow$} & \textbf{Recall $\uparrow$} & \textbf{F1-score $\uparrow$} \\
  \midrule
  
  \multirow{6.5}{*}{\rotatebox{90}{GoodNews}} 
  & LLaVA-1.6-7B~\citep{liu2023visual}     & 1.90 & 9.41 & 12.25 & 15.08 & 15.58 & 12.57 & 13.91 \\  
  & Qwen2.5-VL-7B~\citep{yang2024qwen2_5}  & 2.66 & 8.33 & 12.51 & 20.66 & 17.12 & 11.57 & 13.81 \\ 
  & Qwen2.5-VL-32B~\citep{yang2024qwen2_5} & 3.02 & 11.10 & 14.98 & 12.78 & \underline{24.45} & 20.28 & 22.17 \\ 
  & GPT-4o~\citep{hurst2024gpt}   & 3.58 & \underline{12.80} & \underline{18.00} & 18.88 & 22.58 & \underline{27.01} & \underline{24.60} \\ 
  & Claude-3.5-Sonnet~\citep{anthropic2024claude35sonnet}    & \underline{5.71} & 9.96 & 16.38 & \underline{29.05} & 24.42 & 16.00 & 19.33 \\ 
  \cmidrule{2-9}
  & MERGE & \textbf{10.19}  & \textbf{14.31} & \textbf{28.02} & \textbf{94.54} & \textbf{32.89} & \textbf{31.93} & \textbf{32.40} \\
  \cmidrule{1-9}
  
  \multirow{6.5}{*}{\rotatebox{90}{NYTimes800k}}
  & LLaVA-1.6-7B~\citep{liu2023visual}     & 1.63 & 7.50  & 10.36 & 11.24 & 14.11 & 11.10 & 12.43 \\  
  & Qwen2.5-VL-7B~\citep{yang2024qwen2_5}  & 2.94 & 9.20  & 13.47 & 19.04 & 20.24 & 15.40 & 17.49 \\ 
  & Qwen2.5-VL-32B~\citep{yang2024qwen2_5} & 3.26 & 11.76 & 16.05 & 14.01 & 24.81 & 23.27 & 24.02 \\ 
  & GPT-4o~\citep{hurst2024gpt}   & 3.18 & \underline{13.27} & 16.78 & 9.30 & 23.33 & \underline{30.60} & 26.47 \\
  & Claude-3.5-Sonnet~\citep{anthropic2024claude35sonnet}    & \underline{4.20} & 11.32 & \underline{17.66} & \underline{31.40} & \underline{25.92} & 20.20 & \underline{28.40} \\ 
  \cmidrule{2-9}
  & MERGE & \textbf{11.47} & \textbf{14.94} & \textbf{27.51} & \textbf{88.16} & \textbf{31.87} & \textbf{36.04} & \textbf{33.83} \\
  \cmidrule{1-9}

  \multirow{6.5}{*}{\rotatebox{90}{Visual News}}
  & LLaVA-1.6-7B~\citep{liu2023visual}     & 2.22 & 8.01  & 11.80 & 14.81 & 13.59 & 12.93 & 13.25 \\  
  & Qwen2.5-VL-7B~\citep{yang2024qwen2_5}  & 1.88 & 6.59  & 11.92 & 15.38 & 17.49 & 10.67 & 13.25  \\  
  & Qwen2.5-VL-32B~\citep{yang2024qwen2_5} & 1.00 & 4.58  & 9.12 & 4.53 & 18.51 & 5.81 & 8.84 \\  
  & GPT-4o~\citep{hurst2024gpt}            & \underline{3.97} & \underline{13.89}  & \underline{17.63} & 13.28 & \underline{19.82} & \textbf{30.52} & \textbf{29.99} \\  
  & Claude-3.5-Sonnet~\citep{anthropic2024claude35sonnet} & 2.98 & 11.45  & 15.33 & \underline{13.46} & 18.97 & 23.90 & 21.15 \\  
  \cmidrule{2-9}
  & MERGE & \textbf{14.77}  & \textbf{15.72} & \textbf{28.26} & \textbf{127.77} & \textbf{29.88} & \underline{29.45} & \underline{29.66} \\
  \bottomrule
\end{tabular}}
\caption{Performance comparison with advanced MLLMs directly applied to the news image captioning task. Given budget limitations, we limited the evaluation of GPT-4o and Claude-3.5-Sonnet to 1,000 samples.}
\label{tab:MLLM-results}
\end{table*}

\begin{table*}[ht]
\centering
\small
\renewcommand{\arraystretch}{1.1}
\resizebox{0.99\textwidth}{!}{
\begin{tabular}{ll*{7}{c}} 
  \toprule
  &  & \multicolumn{4}{c}{\textbf{Caption Quality}} & \multicolumn{3}{c}{\textbf{Named Entity Accuracy}} \\ \cmidrule(lr){3-6} \cmidrule(lr){7-9} 
  \multirow{-2.5}{*}{\textbf{Dataset}} & \multirow{-2.5}{*}{\textbf{Method}} & \textbf{BLEU-4 $\uparrow$} & \textbf{METEOR $\uparrow$} & \textbf{ROUGE $\uparrow$} & \textbf{CIDEr $\uparrow$} & \textbf{Precision $\uparrow$} & \textbf{Recall $\uparrow$} & \textbf{F1-score $\uparrow$} \\
  \midrule
  
  \multirow{3}{*}{GoodNews} 
  & MERGE (Qwen2.5-VL-7B) & 8.91 & 13.47 & \underline{27.69} & 89.63 & \textbf{34.05} & 28.36 & 30.94 \\ 
  & MERGE (LLaVA-1.5-7B) & \underline{8.97} & \underline{13.55} & 27.04 & \underline{90.49} & \underline{33.52} & \underline{29.02} & \underline{31.11} \\ 
  & MERGE (InstructBLIP) & \textbf{10.19}  & \textbf{14.31} & \textbf{28.02} & \textbf{94.54} & 32.89 & \textbf{31.93} & \textbf{32.40} \\
  \cmidrule{1-9}
  
  \multirow{3}{*}{NYTimes800k}
  & MERGE (Qwen2.5-VL-7B) & 8.43 & 12.53 & 26.05 & 80.16 & \textbf{35.08} & 28.72 & 31.58 \\ 
  & MERGE (LLaVA-1.5-7B)  & \underline{9.14} & \underline{13.52} & \underline{26.27} & \underline{83.94} & \underline{34.95} & \underline{31.60} & \underline{33.19} \\ 
  & MERGE (InstructBLIP)   & \textbf{11.47}  & \textbf{14.94} & \textbf{27.51} & \textbf{88.16} & 31.87 & \textbf{36.04} & \textbf{33.83} \\
  \cmidrule{1-9}

  \multirow{3}{*}{Visual News}
  & MERGE (Qwen2.5-VL-7B) & \underline{11.58} & \underline{13.22} & \underline{26.79} & \underline{121.29} & \textbf{41.20} & \underline{28.21} & \textbf{33.49} \\ 
  & MERGE (LLaVA-1.5-7B)  & 11.31 & 13.01 & 26.60 & 115.35 & \underline{35.67} & 25.57 & \underline{29.79} \\
  & MERGE (InstructBLIP) & \textbf{14.77}  & \textbf{15.72} & \textbf{28.26} & \textbf{127.77} & 29.88 & \textbf{29.45} & 29.66 \\
  \bottomrule
\end{tabular}}
\caption{Performance comparison of MERGE with different MLLMs as backbones.}
\label{tab:backbone-results}
\end{table*}

\begin{table*}[ht]
\centering
\small
\renewcommand{\arraystretch}{1.1}
\begin{tabular}{l*{4}{c}} 
  \toprule
  & \multicolumn{3}{c}{\textbf{Dataset}} \\ 
  \cmidrule(lr){2-4} 
  \multirow{-2.5}{*}{\textbf{Method}} & \textbf{GoodNews} & \textbf{NYTimes800k} & \textbf{Visual News} & \multirow{-2.5}{*}{\textbf{Average Time $\downarrow$}} \\
  \midrule
  
  \citet{kalarani2023let} & 0.42 & 0.46 & 0.40 & 0.43 \\ 
  EAMA~\citep{zhang2024entity} & 4.12 & 4.04 & 3.79 & 3.98 \\ 
  \citet{xu2024cross} & 1.45 & 2.02 & 1.84 & 1.77 \\ 
  \cmidrule{1-5}
  MERGE(InstructBLIP) & 8.23 & 5.77 & 5.20 & 6.40 \\
  MERGE(Qwen2.5-VL-7B) & 1.61 & 1.87 & 1.48 & 1.65 \\
  \bottomrule
\end{tabular}
\caption{Inference time comparison of different methods (in seconds).}
\label{tab:inferencen-time}
\end{table*}

\section{Implementation Details of EMKB}
\label{sec:impl_emkb}

We detail the practical implementation of the EMKB, which underpins MERGE's ability to enrich captions with external knowledge.

\paragraph{Challenges}
A key challenge in building EMKB is avoiding duplicate or redundant images, especially for lesser-known entities, where distinct visuals are scarce. 
Since many images are crawled from the web, there is also a risk of overlap with the training set, leading to potential data leakage.

\paragraph{Practical Implementation}
To mitigate these issues, we leverage CLIP's image encoder to compute pairwise cosine similarity scores among images. 
Any image pair with a similarity exceeding a predefined threshold $\delta$ is considered redundant. 
This filtering preserves a diverse and high-quality set of entity visuals within the knowledge base.

Specifically, we refine the knowledge base $B$ by retaining only images whose cosine similarity falls below $\delta = 0.95$.
Given an image $\bm{I}_i$ from $\bm{B}$ and an training image $\bm{I}_j$ from $\bm{H}$, the filtered knowledge base $\bm{B}^{\prime}$ is defined as: 
\begin{displaymath}
  \bm{B}^{\prime} = \{\bm{I}_i \mid \bm{I}_i \in \bm{B}, \bm{I}_j \in \bm{H}: \cos(\bm{I}_i, \bm{I}_j) \leq \delta \}. 
\end{displaymath}

This deduplication ensures a robust and diverse entity-aware visual dataset.
Table \ref{table:entity_database} summarizes the statistical distribution of entities and their associated images across different data sources. 
To cover less common or missing celebrity entities, we additionally incorporate data from specialized celebrity datasets such as CACD~\citep{chen2014cross}, IMDb-WIKI~\citep{Rothe2015DEX}, IMDb-Face~\citep{wang2018devil}, and VGGFace2~\citep{Cao2018vggface2}. 

\paragraph{Updating Mechanism}
To ensure EMKB remains current and adaptable to new content, we also implement a dynamic updating mechanism. 
As new entities emerge in news articles or external knowledge sources, MERGE continuously integrates them into EMKB by extracting entity names and associated context using LLM-based entity recognition. 
For each new entity, relevant images are collected from public sources such as Wikipedia and Google Search, followed by deduplication through CLIP-based similarity filtering to maintain diversity and avoid redundancy. 
Additionally, background information for new entities is structured into knowledge subgraphs using the same prompt-based extraction pipeline described in Section~\ref{sec:prompts:knowledge}. 

This modular design enables EMKB to expand seamlessly, ensuring MERGE retains high accuracy and relevance in rapidly evolving news scenarios.

\section{Dataset Preparation}
\label{sec:dataset_details}

For the GoodNews and NYTimes800k datasets, we follow the dataset splits defined by Tell~\citep{tran2020transform}, partitioning GoodNews into 18,335 validation samples and 23,051 test samples, while NYTimes800k comprises 7,777 validation and 21,977 test samples. 
To enhance training efficiency and keep consistency with prior work, such as EAMA~\citep{zhang2024entity}, we randomly sample 100,000 instances from each training set.

For the Visual News dataset, which was not part of EMKB's construction, we adopt a similar approach. 
We randomly sample 100,000 images from each of the four contributing news agencies, forming a combined training set of 400,000 samples.
Following the protocol from \citet{liu2021visual}, we also prepare validation and test sets, each containing 40,000 images.
This setup ensures fair comparisons across datasets while keeping computational demands manageable for large-scale experiments.

\section{Implementation Details of MERGE}
\label{sec:impl_method}

We now present the implementation details of MERGE.
Entity extraction and knowledge graph construction rely on \texttt{Qwen2.5-32b}.\footnote{\url{https://huggingface.co/Qwen/Qwen2.5-32B-Instruct-AWQ}}
For cross-modal alignment, HCMA uses \texttt{InternVL2-Llama3-76B-AWQ}\footnote{\url{https://huggingface.co/OpenGVLab/InternVL2-Llama3-76B-AWQ}} to identify sentences closely tied to visual content.
For visual-entity grounding, RMKI leverages InsightFace for face recognition, specifically the \texttt{buffalo\_l} model, which uses SCRFD-10GF\citep{guo2022sample} for face detection and ResNet50 \citep{he2016deep} trained on WebFace600K for face recognition. 
Faces are detected with a confidence threshold of 0.8, and each detected face is matched to its top-1 nearest neighbor via cosine similarity.
For non-facial images, RMKI utilizes CLIP\footnote{\url{https://huggingface.co/openai/clip-vit-base-patch32}} embeddings to match images to known entities in EMKB.
Caption generation is driven by InstructBLIP~\citep{dai2023instructblip}, further enhanced by a 4-layer GAT~\citep{velickovic2018graph} to encode structured background knowledge, improving context understanding and factual correctness.

InstructBLIP processes input sequences of up to $n_{ctx} = 1,024$ tokens and generates captions with a maximum length of $n_{out} = 50$ tokens.
Training is performed using PyTorch~\citep{paszke2019pytorch} and DeepSpeed~\citep{rasley2020deepspeed} to enable efficient distributed learning.
Fine-tuning of the InstructBLIP decoder is conducted with QLoRA~\citep{dettmers2023qlora}, using a LoRA rank $r=16$ and scaling factor $\alpha=16$. 
The complete model encompasses approximately 7B parameters, with about 466M trainable parameters updated during fine-tuning.

We employ the AdamW optimizer~\citep{loshchilov2019decoupled}, adjusting learning rates for each dataset:
(1) $\eta_1 = 3.0 \times 10^{-5}$ for GoodNews, 
(2) $\eta_2 = 2.0 \times 10^{-5}$ for NYTimes800k, and
(3) $\eta_3 = 2.0 \times 10^{-5}$ for Visual News, 
with a fixed epsilon $\epsilon = 1.0 \times 10^{-4}$.
MERGE is trained for five epochs with a batch size of $B = 16$ and random seed of $s = 42$ across four NVIDIA GeForce RTX 3090 GPUs (24GB VRAM each) with an Intel Xeon CPU and 125GB system memory, running on Ubuntu 18.04 LTS with PyTorch framework.
Table~\ref{table:hyperparameters} details our experimental hyperparameters.

For experiments with alternative MLLM backbones, such as LLaVA-1.5-7B and Qwen2.5-VL-7B, we followed official LoRA fine-tuning protocols, adjusting learning rates to $2.0 \times 10^{-5}$ and expanding the maximum sequence length $n_{seq}$ to 4,096 for the GoodNews,  NYTimes800k, and Visual News datasets.

\section{Detailed NER Results} 
\label{sec:ner_results}

Table \ref{tab:entity-results} presents the detailed NER results, highlighting MERGE's substantial improvements over state-of-the-art baselines across all three entity types.

For \textbf{PERSON} entities, MERGE achieves substantial F1-score improvements: +13.37 on GoodNews and +6.65 on NYTimes800k over the best prior method \citep{xu2024rule}, and +9.66 on Visual News compared to \citet{kalarani2023let} (reproduced using their official code).
While \citet{xu2024rule} leverages news-specific semantic rules to associate named entities with core actions, their method is limited to entities explicitly mentioned in the article and fails when such mentions are absent.
In contrast, MERGE mitigates this limitation by leveraging multimodal retrieval through RMKI and EMKB, enabling accurate recognition of individuals even when they are missing from the textual context.
Additionally, HCMA contributes by improving cross-modal alignment and refining context selection for generating entity-rich captions.

Beyond \textbf{PERSON} entities, MERGE also achieves consistent improvements for other types of entities. 
For \textbf{GPE} (geopolitical entities), MERGE improves F1-scores by +5.03 on GoodNews, +3.91 on NYTimes800k, and +7.12 on Visual News. 
For \textbf{ORG} (organizations), gains reach +5.98, +4.56, and +8.33 on the three datasets, respectively.
These consistent gains across entity categories validate MERGE's ability to integrate structured knowledge retrieval with precise cross-modal alignment for comprehensive named entity recognition.

\section{Evaluation against Advanced MLLMs}
\label{sec:mllms_results}

Table \ref{tab:MLLM-results} compares MERGE against several advanced MLLMs applied directly to the news image captioning task.
The evaluated MLLMs include LLaVA-1.6 \citep{liu2023visual}, Qwen2.5-VL-7B/32B \citep{yang2024qwen2_5}, GPT-4o \citep{hurst2024gpt}, and Claude-3.5 \citep{anthropic2024claude35sonnet}. 
All were evaluated on GoodNews, NYTimes800k, and Visual News, with GPT-4o and Claude-3.5 tested on 1,000 representative samples due to API constraints.

Despite their strong general capabilities, these MLLMs consistently underperform MERGE, particularly on metrics like CIDEr and F1-score that require precise entity recognition and deep contextual integration. 
These results underscore the limitations of generic MLLMs for specialized, knowledge-intensive tasks such as news image captioning.
In contrast, MERGE's domain-tailored architecture--combining structured retrieval, chain-of-thought reasoning, and multimodal alignment--proves significantly more effective, highlighting the critical importance of domain adaptation in specialized applications.

\section{Backbone Versatility of MERGE}
\label{sec:backbones_analysis}

To assess MERGE's flexibility, we evaluated alternative MLLM backbones beyond InstructBLIP.
Specifically, we integrated LLaVA-1.5-7B \citep{liu2023visual} and Qwen2.5-VL-7B \citep{yang2024qwen2_5} into MERGE. 
Results are shown in Table \ref{tab:backbone-results}, where MERGE (MLLM) indicates the specific MLLM backbone in use.

Among these configurations, InstructBLIP consistently delivers the best performance, owing to its Q-Former architecture, which effectively compresses visual tokens and allocates more capacity to contextual and linguistic reasoning.
Importantly, regardless of the backbone, MERGE consistently outperforms prior MLLM-based baselines such as \citet{xu2024cross} (LLaVA) and EAMA (InstructBLIP) \citep{zhang2024entity} across most metrics, as shown in Table 1.
This confirms that MERGE's gains stem from its architectural innovations--including structured retrieval and guided alignment--rather than any single MLLM backbone.

\begin{figure*}[!t]
  \centering
  \includegraphics[width=0.99\textwidth]{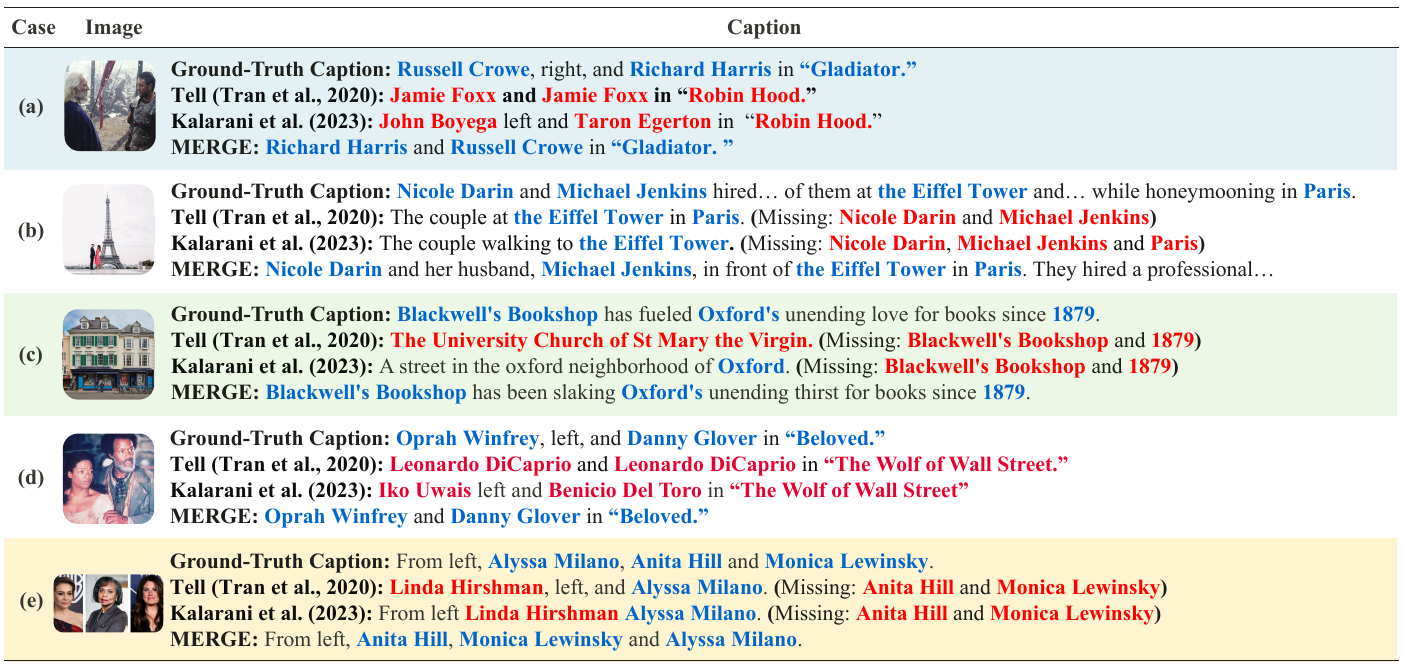}
  \caption{Case study on NYTimes800k. Entities correctly identified by MERGE are depicted in blue, while errors are shown in red.}
  \label{fig:case_study_nytimes}
\end{figure*}

\begin{figure*}[!t]
  \centering
  \includegraphics[width=0.99\textwidth]{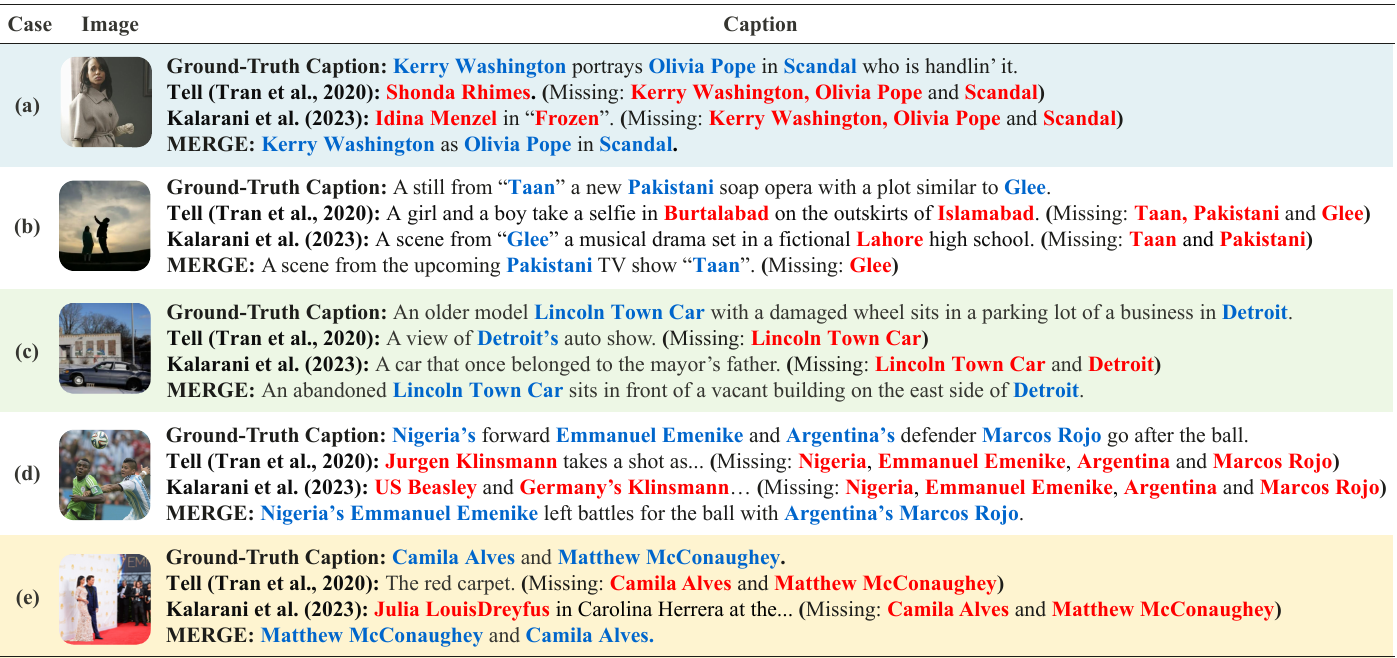}
  \caption{Case study on Visual News. Entities correctly identified by MERGE are depicted in blue, while errors are shown in red.}
  \label{fig:case_study_visualnews}
\end{figure*}

\section{Inference Time and Scalability}
\label{sec:inference_time}

Table~\ref{tab:inferencen-time} presents the inference time for different methods.
In our default setup, MERGE employs InstructBLIP as the MLLM backbone, resulting in an average inference time of 6.40 seconds per instance on GoodNews, NYTimes800k, and Visual News.
While traditional non-LLM methods such as \citet{kalarani2023let} are faster (0.43 seconds), they lack the contextual depth and precision of MERGE.
MLLM-based alternatives like EAMA~\citep{zhang2024entity} and \citet{xu2024cross} offer moderate latency (3.98 seconds and 1.77 seconds, respectively) but still fall short in caption quality and named entity accuracy.

Notably, news image captioning is typically conducted in \emph{offline} workflows, where slightly higher latency is acceptable given MERGE’s superior contextual grounding and accuracy.
To improve runtime efficiency, we also tested Qwen2.5-VL-7B~\citep{yang2024qwen2_5} as a drop-in replacement for InstructBLIP, which reduced average inference time to 1.65 seconds, demonstrating MERGE's flexibility in adopting lightweight MLLMs.

\section{Additional Case Studies}
\label{sec:impl_case}

\paragraph{Case Study on NYTimes800k}
Figure \ref{fig:case_study_nytimes} presents qualitative examples from NYTimes800k, reinforcing the findings from GoodNews (Figure 4). 
Key observations include:
\begin{itemize}
  \item \textbf{Information Enhancement (via EMKB):} 
  In Case (a), MERGE successfully retrieves entities like \texttt{Richard Harris} that are missing from the text, whereas baseline methods fail to generate even explicitly mentioned entities such as \texttt{Russell Crowe}.

  \item \textbf{Fine-grained Cross-modal Alignment (via HCMA):} 
  In Cases (b) and (c), MERGE captures nuanced event and temporal cues (e.g., \texttt{hired a photographer}, \texttt{1879}) that are often missed by prior approaches.
  
  \item \textbf{Precise Visual-Entity Alignment (via RMKI):} 
  Cases (d) and (e) demonstrate MERGE’s robustness in identifying multiple individuals or distinguishing visually similar subjects, where baselines frequently misidentify or omit entities.
\end{itemize}

\paragraph{Case Study on Visual News}
Figure~\ref{fig:case_study_visualnews} showcases case studies from Visual News, which was excluded from EMKB construction. This makes it a rigorous test of MERGE's generalization ability to unseen data domains.
Notable strengths observed:
\begin{itemize}
  \item \textbf{Information Enhancement (via EMKB):} 
  In Case (a), MERGE accurately retrieves missing context, such as \texttt{Kerry Washington}, absent from the text, while baselines fail to identify even clearly mentioned entities like \texttt{Scandal}.

  \item \textbf{Fine-grained Cross-modal Alignment (via HCMA):} 
  Cases (b) and (c) illustrate MERGE's capacity to surface detailed, contextually relevant information (e.g., \texttt{upcoming Pakistani TV show}, \texttt{Lincoln Town Car}) that is consistently overlooked by existing models.
  
  \item \textbf{Precise Visual-Entity Alignment (via RMKI):} 
  In Cases (d) and (e), MERGE accurately identifies multiple entities in complex scenes, outperforming baselines that struggle with disambiguation or omission.
\end{itemize}

\paragraph{Summary}
These cases highlight MERGE's effectiveness in enriching context, aligning modalities, and achieving precise visual-semantic grounding, confirming its effectiveness across diverse datasets.


\end{document}